\documentclass{article}

\usepackage{arxiv}

\usepackage[utf8]{inputenc} 
\usepackage[T1]{fontenc}    
\usepackage{hyperref}       
\usepackage{url}            
\usepackage{booktabs}       
\usepackage{amsfonts}       
\usepackage{nicefrac}       
\usepackage{microtype}      
\usepackage{lipsum}

\usepackage{graphicx}
\usepackage{amsmath,amssymb}

\newcommand{\beginsupplement}{%
        \setcounter{table}{0}
        \renewcommand{\thetable}{S\arabic{table}}%
        \setcounter{figure}{0}
        \renewcommand{\thefigure}{S\arabic{figure}}%
     }

\title{Discovering a sparse set of pairwise discriminating features in high dimensional data}

\author{
  Samuel Melton\\
 John A. Paulson School of Engineering and Applied Sciences \\
 Harvard University\\
 Cambridge, MA, 02138 \\
 \texttt{smelton@g.harvard.edu}  \\
   \And
 Sharad Ramanathan \\
 John A. Paulson School of Engineering and Applied Sciences\\
 Department of Stem Cell and Regenerative Biology\\
 Department of Molecular and Cellular Biology\\
 Harvard University \\
 Cambridge, MA, 02138
}

\begin{document}
\maketitle

\begin{abstract}
Extracting an understanding of the underlying system from high dimensional data is a growing problem in science. Discovering informative and meaningful features is crucial for clustering, classification, and low dimensional data embedding. Here we propose to construct features based on their ability to discriminate between clusters of the data points. We define a class of problems in which linear separability of clusters is hidden in a low dimensional space. We propose an unsupervised method to identify the subset of features that define a low dimensional subspace in which clustering can be conducted. This is achieved by averaging over discriminators trained on an ensemble of proposed cluster configurations. We then apply our method to single cell RNA-seq data from mouse gastrulation, and identify 27 key transcription factors (out of 409 total), 18 of which are known to define cell states through their expression levels. In this inferred subspace, we find clear signatures of known cell types that eluded classification prior to discovery of the correct low dimensional subspace.
\end{abstract}


\section{Introduction}

Recent technological advances have resulted in a wealth of high dimensional data in biology, medicine, and the social sciences. In unsupervised contexts where the data is unlabeled, finding useful representations is a key step towards visualization, clustering, and building mechanistic models. Finding features which capture the informative structure in the data has been hard,  however, both because of unavoidably low data density in high dimensions (the ``curse of dimensionality" \cite{Donoho00high-dimensionaldata}) and because of the possibility that a small but unknown fraction of the measured features define the relevant structure (e.g. cluster identity) while the remaining features are uninformative \cite{skmeans,pca_clust}.

Identifying informative features has long been of interest in the statistical literature. When the data is labeled, allowing for a supervised analysis, there are successful techniques for extracting important features using high dimensional regressions. When there is no labeled training data, unsupervised discovery of features is difficult. Standard feature extraction methods such as PCA are effective in reducing dimensionality, yet do not necessarily capture the relevant variation \cite{pca_clust} (and Supplemental Figure 1). Other methods attempt \cite{skmeans,max_margin_clust} to co-optimize a cost function depending on both cluster assignments and feature weights, which is computationally difficult and tied to specific clustering algorithms (See section on existing methods). Feature extraction guided by clustering has also been effective as a preprocessing step for regression tasks. In \cite{Coates2012}, classification and regression is done with data represented in the basis of centroids found with K-Means. In \cite{sparse_filtering}, features are constructed such that representations of data points are sparse, but no explicit discrimination is encoded between clusters beyond a sparsity constraint. We consider here an example where clusters are distinguished from each other by sparse features, but overall representations of each data point is not necessarily sparse in this new basis. We show here that optimal features are discovered by their ability to separate pairs of clusters, and we find them by averaging over proposed clustering configurations.

Using gene expression data to understand processes in developmental biology highlights this challenge. In a developing embryo, multi-potent cells make a sequence of decisions between different cell fates, eventually giving rise to all the differentiated cell types of the organism. The goal is both to determine the physiological and molecular features that define the diversity of cell states, and to uncover the molecular mechanisms that govern the generation of these states. Decades of challenging experimental work in developmental biology suggests that a small fractions of genes control specific cell fate decisions \cite{Graf_2009,gilbert2016developmental,Takahashi_2006}. Recent experimental techniques measure tens of thousands of features -- gene expression levels -- from individual cells obtained from an embryo over the course of development, producing high dimensional data sets \cite{Farrelleaar3131,Briggs:2018aa}. Clustering these data to extract cell states and identifying the small fractions of key genes that govern the generation of cellular diversity during development has been difficult \cite{Kiselev_2019, Furchtgott_2017}. However, mapping cellular diversity back to specific molecular elements is a crucial step towards understanding how gene expression dynamics lead to the development of an embryo.

Here we show that as the fraction of relevant features decreases, existing clustering and dimensionality reduction techniques fail to discover the identity of relevant features. We show that when the linear separability of clusters is restricted to a subspace, the identity of the subspace can be found without knowing the correct clusters by averaging over discriminators trained on an ensemble of proposed clustering configurations. We then apply it to previously published single-cell RNA-seq data from the early developing mouse embryo \cite{gottgens}, and discover a subspace of genes in which a greater diversity of cell types can be inferred. Further, the relevant subspace of genes that we discover not only cluster the data but are known from the experimental literature to be instrumental in the the generation of the different cell types that arise at this stage. This approach provides unsupervised sparse feature detection to further mechanistic understanding and can be broadly applied in unsupervised data analysis.

\section{Results}

\subsection{Uninformative Data Dimensions Corrupt Data Analysis}

\begin{figure}
\includegraphics[width=.6\linewidth]{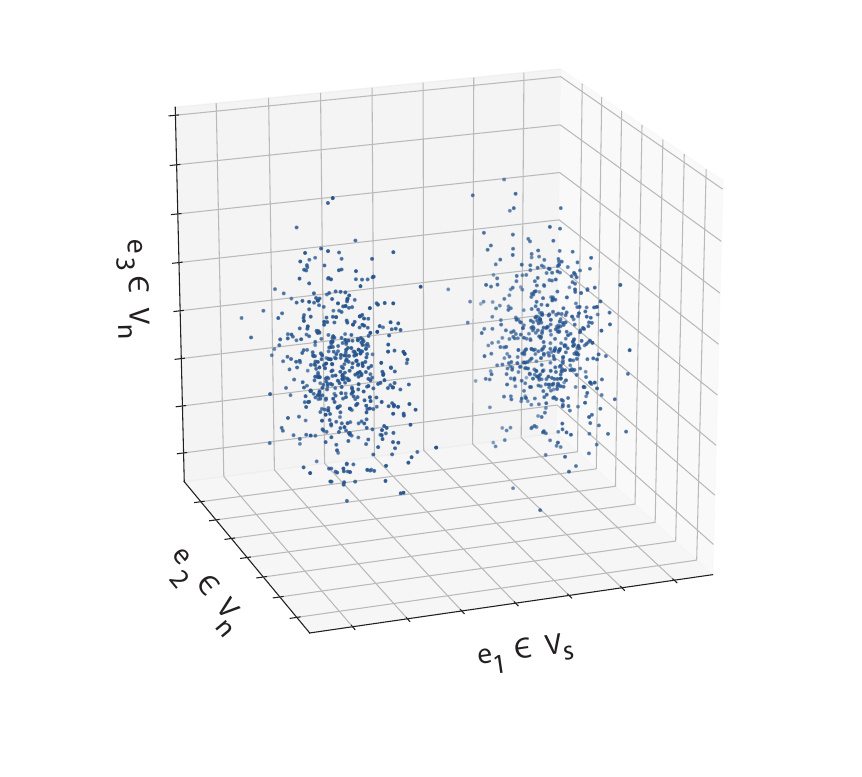}
\centering
\caption{Gaussian data with unit variance shown along 3 axes. The marginal distribution of $e_1$ contains signature of distinct clusters, with a bimodal marginal distribution where each mode corresponds to a cluster. Here the clusters are linearly separable along the $e_1$ axis. The marginal distributions of $e_2$ and $e_3$ are unimodal, and do not linearly separate groups of data points. Here we designate $e_1$ as part of $V_s$ as it contains multimodal signal, and $e_2,e_3 \in V_n$ do not.}
\label{fig1} 
\end{figure}

To understand how the decreasing fraction of relevant features affects data analysis, consider data from $K^{true}$ classes in a space $V$ with $dim(V)=D$ features.  Assume that $V$ can be partitioned into two subspaces. First, an informative subspace $V_s$ of dimension $D_s$, in which the $K^{true}$ clusters are separable. And second, an uninformative subspace $V_n$ with dimension $D_n = D - D_s$ in which the $K^{true}$ clusters are not separable. An example of such a distribution is shown in Fig. \ref{fig1} with two clusters, $D_s = 1$ and $D_n = 2$. 

The correlation between the distances computed in the full space $V$ with that in the relevant subspace $V_s$ scales as $\sqrt{D_s/D}$ (see Supplemental Text). When the fraction of relevant features is small, or equivalently $D/D_s \gg 1$, correlations between samples become dominated by noise. In this regime, without the correct identification of $V_s$, unsupervised analysis of the data is difficult, and typical dimensionality reduction techniques (PCA, ICA, UMAP, etc) fail. We demonstrate this by constructing a Gaussian mixture model with 7 true clusters which are linearly separable in a subspace $V_s$ with dimension $D_s = 21$, and drawn from the same distribution (thus not linearly separable) in the remaining $D-D_s$ dimensions. As the ratio $D /D_s$ increases, the separability of the clusters in various projections decreases (Supplemental Fig. 1). 

In many cases, identifying the ``true" $V_s$ may be challenging. However, eliminating a fraction of the uninformative features and moving to a regime of smaller $D/D_s$ could allow for more accurate analysis using classical methods. We next outline a method to weight dimensions to construct an estimate of $V_s$ and to reduce $D/D_s$.

\subsection{Minimally Informative Features: The Limit of Pairwise Informative Sub-spaces Separating Clusters}

To develop a framework to identify the relevant features, consider data $\textbf{X} = \{ \vec{x}_1,..,\vec{x}_N \}$ where samples $\vec{x}_i$ are represented in the measurement basis $\{ \vec{e}_1, ..., \vec{e}_D \}$. Assume that the data is structured such that each data point is a member of one of $K^{true}$ clusters, $\mathcal{C} \equiv \{C_1,..,C_{K^{true}}\}$. Let $V^{lm}_s$ be the subspace of $V$ in which the data points belonging to the pair of clusters $C_l$, $C_m$ are linearly separable. Let $\vec{\theta}_{lm}$ be unit vector normal to the max-margin hyperplane separating clusters $C_l$ and $C_m$. In the space orthogonally complement to $V^{lm}_s$, the two clusters are not linearly separable. One can similarly define $K^{true}(K^{true}-1)/2$ such subspaces $\{V_s^{lm}\}$ and associated hyperplanes $\{ \vec{\theta}_{lm}\}$, one for each pair of clusters in $\mathcal{C}$. We define a weight for each dimension $\vec{g} = \{g_1,...,g_d,...,g_D \}$ by it's component on the $\{ \vec{\theta}_{lm} \}$s: 

 \begin{equation} \label{g_true}
 g_d ( \{ \vec{\theta}_{lm} \}) =  \sum_{l \neq m} \left|  \vec{\theta}_{lm} \cdot  \vec{e}_d \right|
 \end{equation}
 
Knowing the cluster configuration $\mathcal{C}$ would allow us to directly compute $\vec{g}$ by finding max-margin classifiers and using Equation \ref{g_true}. Conversely, knowing $\vec{g}$ would allow for better inference of the cluster configuration because restriction to a subspace in which $g_d > 0$ would move to a regime of smaller $D/D_s$. Existing work has focused on finding $\mathcal{C}$ and $\vec{g}$ simultaneously, through either generative models or optimizing a joint cost function. Such methods either rely on context specific forward models, or tend to have problems with convergence on real data sets (see ref. \cite{skmeans} and section on existing methods). 

\begin{figure*}
\centering
\includegraphics[width=13cm]{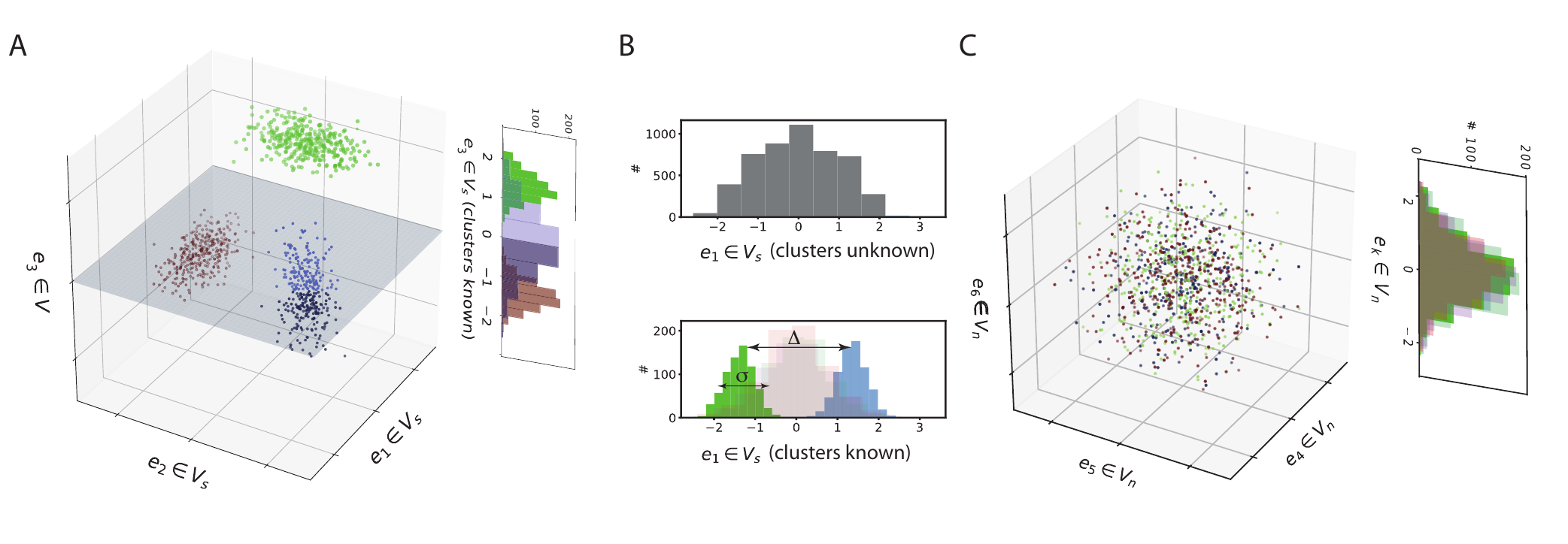}

\centering
\caption{For $K$ clusters with multimodal subspaces $V_s^{lm}$ with $l,m \in \{1,...,K\}$, we consider the limit as each $V_s^{lm}$ has minimal dimension ( = 1) and are non intersecting. A) shows a Gaussian example of a collection of one-dimensional pairwise informative subspaces, which are uninformative for clusters $\neq l,m$. Here $e_1$ is multimodal in the blue and green clusters, but not red, $e_2$ is multimodal in the red and blue clusters, but not green, and $e_3$ is multimodal in the red and green clusters, but not blue. B) Despite containing multimodal signature, non-intersecting pairwise informative subspaces $V_s^{lm}$ can corrupt marginal distributions to hide separability (top). Same data with points colored by cluster, where separation of means is denoted by $\Delta$, and the variance of distributions in their informative dimensions is given by $\sigma$. (bottom). C) shows dimensions that are uninformative for all clusters.}

\label{fig2} 
\end{figure*}

We focus here on estimating $\vec{g}$ when $\mathcal{C}$ is unknown. We consider the limit in which the dimensions of each $V_s^{lm}$, $D_s^{lm}$ take on the smallest possible value of $1$, which maximizes the ratio $D/D_s^{lm}$ for all $l, m$. Further, this limit resides in the regime of large $D/D_s$ where conventional methods fail. We further consider the limit where the intersection between any pair of the subspaces in $\{V_s^{lm}\}$ is null. In this limit, the marginal distribution of all of the data in any one of the $V_s^{lm}$ can be appear unimodal due to a dominance of data points from the $K^{true}-2$ clusters for which this subspace is irrelevant, despite data in the clusters $C_l$, $C_m$ showing a bimodal signature in this subspace. Hence finding the identity of the informative subspaces by distinguishing moments of the marginal distribution is not possible as $D/D_s$ grows or data density decreases, even in the case of normally distributed data (Fig. \ref{fig2}). In this limit, the values of $g_d$ corresponding to informative dimensions are $1/\binom{K^{true}}{2}$, and 0 for uninformative dimensions. Our reason for studying this limit of pairwise separability is that an algorithm that can find the informative subspaces of $V$ in these limits should be able to do so in instances where in the dimensions of $\{V_s^{lm}\}$ are larger than one and intersecting. 

We generate data such that the mean of the marginal distributions of clusters $C_l$ and $C_m$ along a specific $\vec{e}_d$ whose span defines $\{V_s^{lm}\}$ are separated by $\Delta$ and the sample variance of each cluster's marginal distribution is $\sigma$. The marginal distribution of cluster $C_l$ in all other dimensions, i.e. $\vec{e}_d \notin \cup_a V_s^{la}$, is unimodal with zero mean and unit variance. Therefore, in all there are $D_s = K^{true}(K^{true}-1)/2$ dimensions in each of which which a pair of clusters are linearly separable, while the other $K^{true}-2$ clusters are not, and $D_n= D-D_s$ dimensions where all clusters are drawn from the same unimodal distribution. Normalizing each feature to have unit variance leaves one free parameter, $S = \Delta/\sigma$, which controls the pairwise separability of clusters within their informative subspace (Fig. \ref{fig2}B). Indeed, computing pairwise distances between data points generated from 7 clusters and $D/D_s = 40$ does not reveal cluster identity (Fig. \ref{fig3}A).

\subsection{Identifying a Sparse Set of Pairwise Informative Features}

We develop an approach to estimate the weight vector $\vec{g}$ knowing neither the identity of points belonging to each cluster nor the total number of clusters. In order to estimate $\vec{g}$, we propose to average estimates of $\vec{g}$ over an ensemble of clustering configurations. Specifically, we sample an ensemble of possible clustering geometries, $\mathcal{C}^p$, from each of which a collection of max-margin classifiers $\{ \vec{\theta}_{lm}^p \}$ are computed in order to compute $\vec{g}$ using Equation \ref{g_true}: 

\begin{equation} \label{g_d_1}
\langle g_d \rangle =  \sum_{\mathcal{C}^p} g_d (\{ \vec{\theta}_{lm}^p \}) P(\mathcal{C}^p|\textbf{X}) \end{equation}
 
Where $P(\mathcal{C}^p | \textbf{X})$ is the probability of a clustering configuration given the data. This sum can be approximated numerically through a sampling procedure, where cluster proposals are sampled according to

\begin{equation} \label{clust_prob} P (\mathcal{C} | \textbf{X}) \sim \sum_{K^p} P (\mathcal{C} | \textbf{X},K^{p})P(K^{p}) \end{equation}
where, $K^{p}$ is the number of clusters, and $P(K^{p})$ is our prior over the number of proposal clusters. 

Consider one such proposed clustering configuration with $K^p$ clusters, denoted by $ \mathcal{C}^p =  \{C^p_1,..,C^p_{K^p}\}$ where each $C^p_l$ indexes the data points that belong to the $l$th proposed cluster. For this proposed clustering configuration, we compute a set of $\binom{K^p}{2}$ classifiers that separate each pair of clusters. Based on the assumption that $\vec{g}$ is sparse, or equivalently that the true $\{V_{lm}^s \}$ are low dimensional, we impose an L1-regularized max margin classifier to compute $\{ \vec{\theta}_{lm} \}$ from the data $\textbf{X}$ and the proposed cluster configuration $ \mathcal{C}^p$ as in \cite{Zhu:2003:SVM}: 

\begin{equation}  \label{thetaeq}
\begin{aligned}
\vec{\theta}^p_{lm} = \arg\min_{\vec{\theta}}  \Big[& \sum_{i \in C^p_l} \left[ 1 - \left( \vec{\theta} \cdot \vec{x}_i \right) \right]_{+} \\
+ &\sum_{i \in C^p_m} \left[ 1 + \left( \vec{\theta} \cdot \vec{x}_i \right) \right]_{+}  + \lambda || \vec{\theta} ||_1 \Big]
\end{aligned}
\end{equation}

Where $[\cdot ]_+$ indicates the positive component, and $\lambda$ is a sparsity parameter. We set $\lambda$ such that the expected number of non-zero components in each $\theta_{lm}^p$ is 1. Specifically, we sample $T$ cluster configurations by clustering on random subsets of the data, and average the weights of max-margin classifiers over this ensemble:
  \begin{equation} \label{main_eqn_end} 
  \langle g_d \rangle = \frac{1}{T} \sum_{\mathcal{C} ^p}  \left( \sum_{l< m} \vec{\theta}^p_{lm} \cdot \vec{e}_d \right),
  \end{equation}
This procedure can be carried out explicitly as follows: \\

$\textbf{X} \in \mathbb{R}^{D x N}$ ($N$ instances in $D$ dimensions). \\
For $t< T$:
\begin{enumerate}
\item Pick $n_{subsample}$ points from $\textbf{X} \to \textbf{X}^s$
\item Sample $K_{p} \sim Unif(2,K_{max})$
\item $\mathcal{C}^p \gets $ Cluster $\textbf{X}^s$ into $K_{p}$ clusters
\item For $l < m \in \{0,..,K_{p}\}$:
\subitem $$\vec{\theta}^p_{lm} = \arg\min_{\vec{\theta}} \sum_{i \in C^p_l} \left[ 1 - \left( \vec{\theta} \cdot \vec{x}_i \right) \right]_{+} $$  $$+ \sum_{i \in C^p_m} \left[ 1 + \left( \vec{\theta} \cdot \vec{x}_i \right) \right]_{+}  + \lambda || \vec{\theta} ||_1$$
\item For $d<D$
\subitem $$g_d \gets g_d +  \sum_{l \neq m} \left[  \vec{\theta}^p_{lm} \cdot  \vec{e}_d \right]$$
\item Return $\vec{g}$
\end{enumerate}

\begin{figure*}
\centering
\includegraphics[width=14cm]{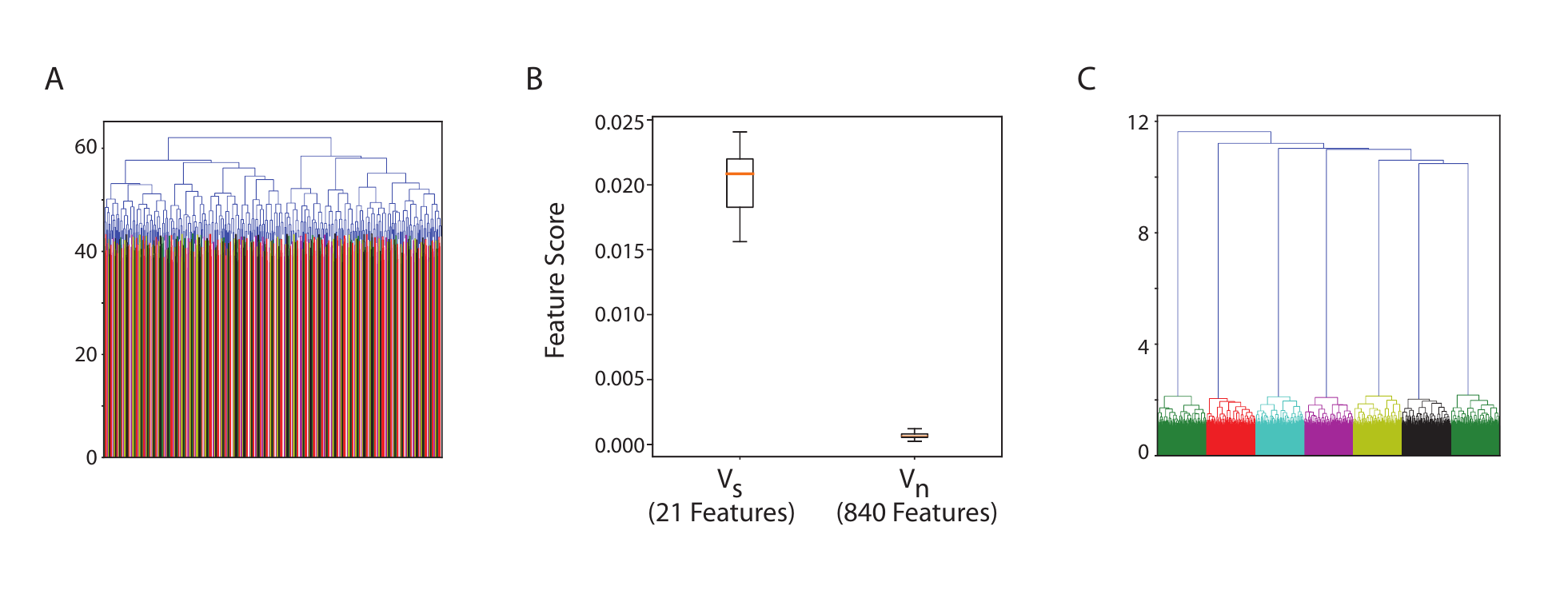}
\centering
\caption{Gaussian data was generated in which 1400 data points from 7 clusters are pairwise distinguishable in only one feature, and 840 features contain no information as to cluster identity (thus $D/D_s = 40$). A) Computing pairwise distances between points and constructing a dendrogram does not resolve the existence of clusters. B) Ensemble of 1000 proposal clusters are constructed using K-Means, with $K_p \sim Unif(3,14)$, and max-margin classifiers are constructed for each pair of cluster per proposal. Each feature is scored according to how frequently it separates two proposed clusters. A histogram of the scores of each feature are shown. Features in the informative subspace ($V_s$), have substantially higher scores than those in the uninformative subspace ($V_n$). C) A dendrogram computed in the space weighted by feature scores reveals the existence of 7 clusters.}
\label{fig3} 
\end{figure*}

While computing pairwise distances in the full space $V$ lacks structure (Fig. \ref{fig3}A), this algorithm produces substantially higher weights for the informative features on simulated data (Fig. \ref{fig3}B). Comparisons of pairwise distances in the reduced subspace found by the algorithm reveals richer structure and the presence of 7 distinct clusters (Fig. \ref{fig3}C). The algorithm reliably discovers the correct set of informative features while using both K-Means and Hierarchical clustering to construct the proposal clusters, and for a range of the prior over $K_p$(Supplemental Fig. 2).

\subsection{Scaling of Inferred Weights with Dimensionality and Data Density}

In the challenging regime of large $D/D_s$, this algorithm can robustly identify key features in the data. In particular, as $D$ increases, there is a scaling of the algorithms performance as a function of $D/D_s$, as well as a dependence on the number of data points $N$. First, we sample a variety of proposal clusters $\mathcal{C}^p$, each with $K^{p}$ clusters drawn from a prior $P(K)$. Using counting arguments (see Supplemental Text) we can estimate the frequency of proposed $\vec{\theta}_{lm}$ aligning with informative with a bimodal signature and uninformative features without. This ratio of the average weights of informative dimensions to the average of the uninformative dimensions, $\langle g_d \rangle _{d \in V_s} / \langle g_d \rangle_{d \notin V_s}$, scales as $\frac{D}{\sqrt{D_s}}$. The scaling, however, also depends on data density. Specifically, consider the length scale separating two neighboring data points in the full space $V$ scales as $N^{-1/D}$. In the relevant subspace $V_s$, this length scale translates to a volume of $N^{-D_s /D}$ which must be compared to the characteristic volumes in this subspace that reflect the multi-modal structure of the data. If the identities of the true clusters in $V_s$ are known, one can ask what the errors are in clustering in the full space $V$ instead of in $V_s$ by computing the entropy, $S$ of the composition of inferred clusters based on the true cluster identities of data points. This entropy has to be a function of the ratio of the characteristic volumes in $D_s$ to $N^{-D_s /D}$. Or equivalently, the entropy of the clusters should be a monotonically increasing function $F(\frac{D}{D_s \log (N)} )$, denoting increasing errors in clustering. The form of the function $F$ depends on the true data distribution and the clustering method. Therefore, our expectation for the ratio of counts for the informative dimensions and counts for the uninformative dimensions should scale like

 \begin{equation} \label{gdgs} \frac{\langle g_d \rangle _{d \in V_s}} {\langle g_d \rangle_{d \notin V_s}} = \frac{D}{\sqrt{D_s}} F\left( \frac{D}{D_s \log (N)} \right) \end{equation}

We numerically generated Gaussian distributed data for $D/D_s \in [2,50]$, $N \in [10^2,10^4]$ using $K_{true} = 7$, $D_s = 21$, and ran 2000 iterations of the algorithm with $P(K^p) \sim \text{Unif}(3,N/20)$ and found close agreement for the range of parameters (Fig. \ref{fig4}A,B). 

\subsection{Significance and Sources of Error}

In order to estimate the significance of $g_d$ frequencies produced by the algorithm, we constructed a null models to estimate $g_d$ in the absence of signal. First, each column of the data matrix is shuffled to produce a null distribution $\textbf{X} \to \textbf{X}^s$. This leaves marginal distributions of each dimension unchanged. Next, each feature can be scored based on the null distribution, $\{g_d^s\}$, and statistics $\mu^s = \langle g_d^s \rangle$, and $\sigma^s = \sqrt{\langle g_d^s \rangle^2 - \langle g_d^{s2} \rangle }$ can be computed from these scores. We then can compute a Z-score for each dimension as $\frac{g_d - \mu^s}{\sigma^s}$. Motivated by the work of \cite{gapstat} and \cite{modelX}, more precise estimates could be obtained by generating synthetic marginals without multimodality, which is an area for future work. 
 
\begin{figure}
\includegraphics[width=.8\linewidth]{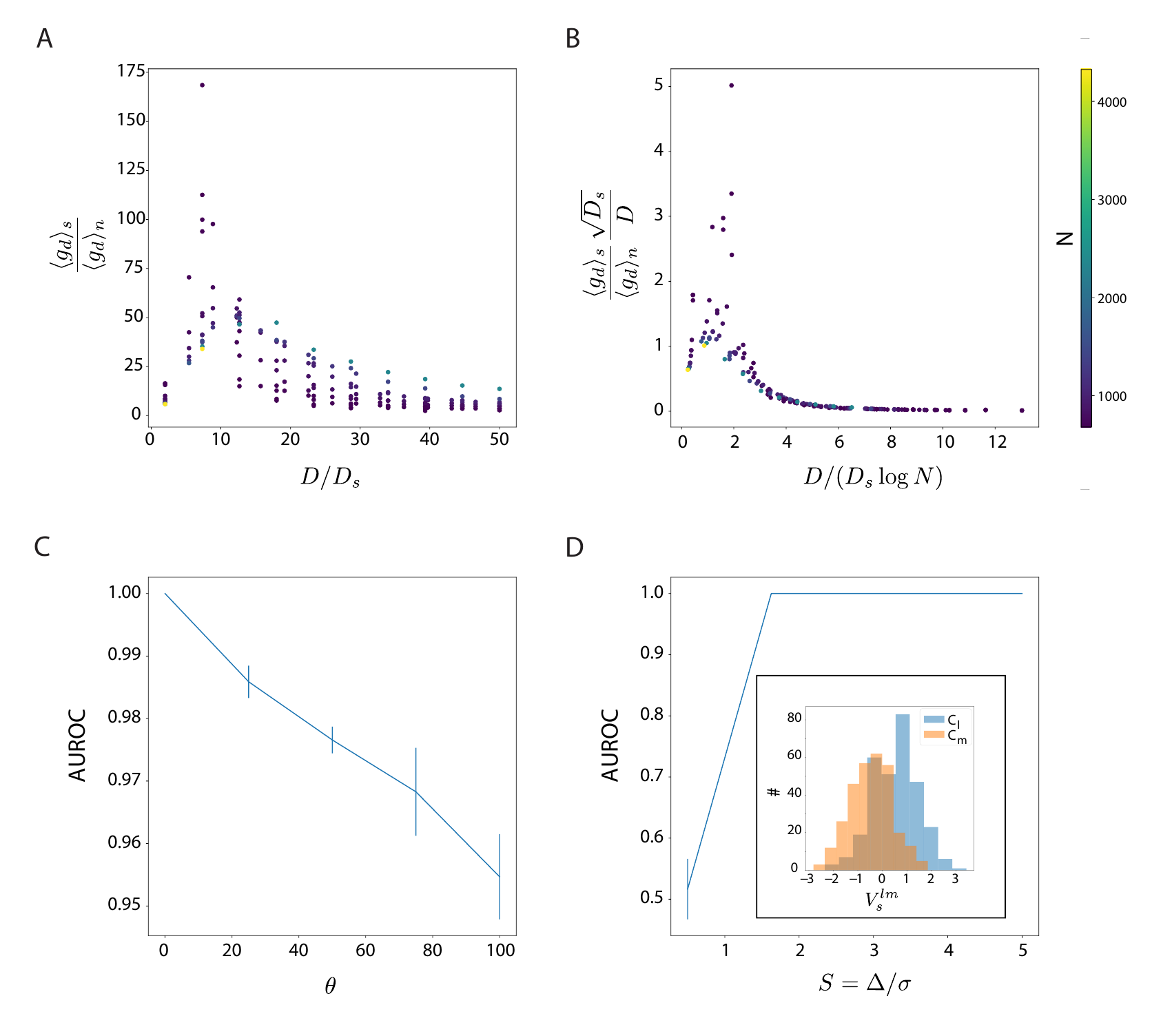}
\centering
\caption {A) We numerically generated Gaussian distributed data for $D/D_s \in [2,50]$, $N \in [10^2,10^4]$ using $K_{true} = 7$, $D_s = 21$, and ran 2000 iterations of the algorithm with $P(K^p) \sim \text{Unif}(3,N/20)$ and the proposal clusters inferred by standard K-means.  B) We find that by scaling by $D / D_s LogN$, we see a consistent trend across number data points and the ratio of counts on informative dimensions to uninformative dimensions matches the predicted $ \frac{D}{\sqrt{D_s}}$ scaling. For larger values of $D / D_s LogN$, the points collapse onto one trend line for various values of $N$. C) Correlations in uninformative dimensions are generated by a i.i.d. standard normal vector $\vec{r}$ is generated, and then the covariance matrix $\Sigma = I + \theta \vec{r} \cdot \vec{r}^T$ where $\theta$ controls the strength of the correlations. As $\theta$ increases, the AUROC decreases, but false negative rates remain low. D)  ROC curves for $K_{true} = 7$, $D/D_sLogN = 15$ as a function of $S = \Delta/\sigma$. Performance deteriorates as separation of clusters in their informative subspaces is reduced to within noise (inset).}

\label{fig4}
\end{figure}

Correlations in the uninformative subspace can lead to erroneous counts in the correlated axes. This is caused by correlations in uninformative dimensions biasing the proposal clusters to be differentially localized in these axes. Despite these false positives, the false negative rate remains low, resulting in minimal degradation of the ROC curve (Fig. \ref{fig4}C). In practice, eliminating any number of uninformative dimensions is effective in restricting analysis to a smaller regime of $D/D_s$. Thus, even in the presence of false positives, removing uninformative dimensions prior to conventional analysis can increase the accuracy of clustering or dimensionality reduction techniques.

A free parameter in the synthetic data generated in Fig. \ref{fig4} is $S=\Delta/\sigma$, the ratio of mean separation to variance of distributions in the informative subspace, which controls the separability of clusters. As $S$ decreases, we see degradation in the AUROC for our algorithm, but identification of key dimensions is still possible even as the mean separation approaches the noise level in the distributions (Fig. \ref{fig4}D, inset).

\subsection{Application to single cell RNA-sequencing from early Mouse Development}

A central challenge in developmental biology is the characterization of cell types that arise during the course of development, and an understanding of the genes which define and control the identity of cells as they transition between states. Starting at fertilization, embryonic cells undergo rapid proliferation and growth \cite{gilbert2016developmental,baldock2015kaufmans}. In a mouse, these cells form the epiblast, a cup shaped tissue surrounded by extra-embryonic cells by E6, or 6 days after fertilization. Only the cells of the epiblast will go on to give rise to all the cells of the mouse. These cells are pluripotent, meaning they have the developmental potential to become any cell type in the adult mouse body \cite{Rossant_2017}. At E6, proximal and distal sub-populations of both the epiblast and surround extra-embryonic cell types begin secreting signaling proteins \cite{Rivera-Perez:2014aa}, which when detected by nearby cells, can increase or decrease the expression of transcription factors -- proteins that modulate gene expression, and can thus change the overall expression profile of a cell. Signaling factors direct genetic programs within cells to restrict their lineage potential and undergo transcriptional as well as physical changes. Posterior - proximal epiblast cells migrate towards the outside of the embryo forming a population called the primitive streak, in a process called gastrulation which takes place between E6.5 and E8. This time frame is notably marked by the emergence of three populations of specified progenitors known as the germ layers \cite{Tam:1997aa}: endoderm cells, which later differentiate into the gastrointestinal tract and connected organs, mesoderm cells, which have the potential to form internal organs such as the muscoskeletal system, the heart, and hematopoietic system, and ectoderm cells, which later form the skin and nervous system. The mesoderm can be subdivided into the intermediate mesoderm, paraxial mesoderm, and lateral plate mesoderm, which each have further restricted lineage potential. Identifying the key transcription factors that define and control the genetic programs that lead to these distinct subpopulations will allow for experimental interrogation and a greater understanding of the gene regulatory networks which control development.

Recent advances in single cell RNA-sequencing technology allow for simultaneous measurement of tens of thousands of genes \cite{Farrelleaar3131,Briggs:2018aa} during multiple time points during development. These technological advances promise to provide insight into the identity and dynamics of key genes that guide the developmental process, yet even clustering cells into types of distinct developmental potential, and identifying the genes responsible for the diversity has been difficult \cite{SPRING,Gr_n_2015_2} . Existing methods typically find signal in correlations between large numbers of genes with large coefficients of variation to determine a cell's states. However, experimental evidence suggests that perturbations of a small number of transcription factors are sufficient to alter a cell's developmental state and trajectory \cite{Graf_2009,gilbert2016developmental,Takahashi_2006}. Further, recent work suggests that a small set of four to five key transcription factors is sufficient to encode each lineage decision \cite{Furchtgott_2017,Petkova_2019}. We therefore believe that signature of structure in these data resides in a low dimensional subspace. While many existing methods rely on hand-picking known transcription factors responsible for developmental transitions \cite{gottgens}, we attempt to discover a low dimensional subspace of gene expression which encodes multi-modal expression patterns indicating the existence of distinct cell states. 

\begin{figure}
\includegraphics[width=15cm]{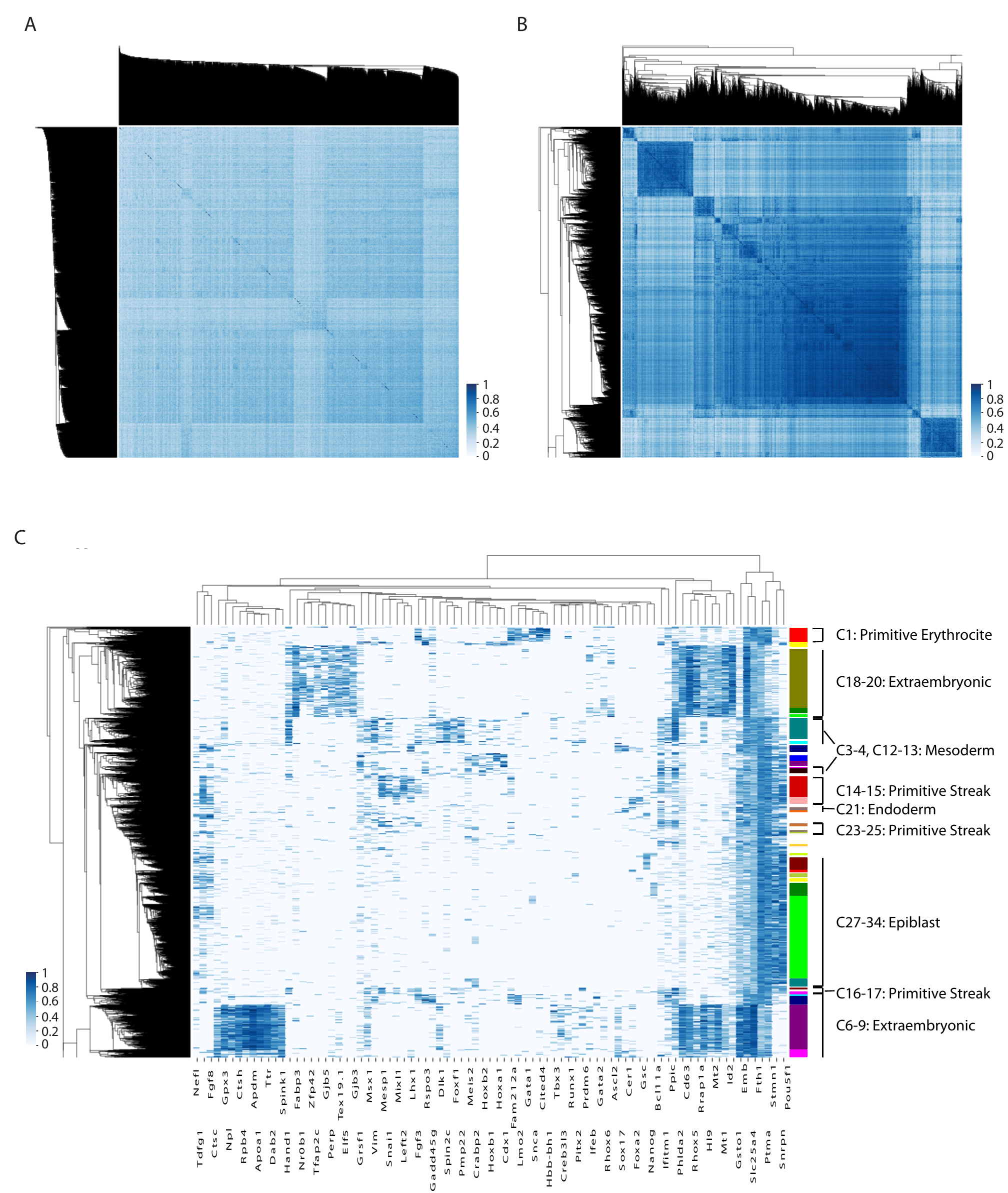}
\centering
\caption{A) Single cell RNA-seq data from \cite{gottgens} does not immediately segregate into cell types. Analysis for A)-C) was conducted on all 48692 cells from E6.5 - E7.75 with at least 10,000 mapped reads, however only 4000 randomly selected cells are shown for visualization purposes. Here we show the cell-cell correlation matrix where each row/column corresponds to a single cell, organized by hierarchical clustering, and the correlation in computed in the 409 dimensional space of expressed transcription factors. B) Inference of 27 transcription factors with pairwise multimodal signature provides a subspace in which to re-compute cell-cell correlations, revealing population structure in comparison to A). C) Inferred transcription factors include known regulators of development and lineage transitions, allowing identification of previously hidden cell types and subpopulations. Here we show normalized expression of inferred transcriptions and correlated genes (columns) vs. single cells (rows) which were clustered hierarchically in this subspace. Differential expression of small numbers of genes distinguishes cell types, such as differential expression of Nanog in C27-C34.}

\label{fig5} 

\end{figure}

In \cite{gottgens}, single-cells are collected from a mouse embryo between E6.5 and E8.5, encompassing the entirety of gastrulation, and profiled with RNA-sequencing to quantify RNA transcriptional abundance. We considered 48692 cells from E6.5 - E7.75 which had more than 10,000 reads mapped to them. We then sub-sampled reads such that each cell had 10,000 reads. Individual genes were removed from analysis if they had a mean value of less than 0.05, or a standard deviation of less than 0.05 (based on \cite{SPRING,Gr_n_2015_2}). We restricted our analysis to transcription factors because, as regulators of other genes, variation in transcription factor expression is a strong indication of biological diversity between cells, or cell types. We normalized the 409 transcription factors with expression above these thresholds to have unit variance. A cell-cell correlation analysis, followed by hierarchical clustering fails to capture the fine grained diversity of cell types that is known to exist at this time point (Fig. \ref{fig5}A).

\begin{tabular}{ p{2cm}|p{1cm}|p{6cm}|p{1.5cm}  }

 \multicolumn{4}{c}{Table 1} \\
 \hline
Gene Name & $z_g$ &  Associated Cell Type & Citation\\
\hline
Creb3l3 &  34.20 & & \\
Tfeb& 13.37  & &\\
Rhox6 & 10.15  & &\\
Elf5 & 9.45  &  Extraembryonic Ectoderm & \cite{Latos_2015} \\
Gata1  & 8.77 & Primitive Erythrocite & \cite{gata1} \\
Pou5f1 & 8.38  & Epiblast, Primitive Streak  &\cite{Mulas:2018aa} \\
Sox17 & 5.61 & Endoderm & \cite{Viotti:2014aa} \\
Nr0b1 & 5.58 & & \\
Hoxb1 & 4.54  & Mesoderm  & \cite{hox} \\
Foxf1 & 3.36 & Lateral Plate Mesoderm & \cite{Mahlapuu:2001aa} \\
Gata2 & 3.27  & Extraembryonic Mesoderm & \cite{gata2} \\
Prdm6 & 3.12 & & \\
Bcl11a & 2.99  & &\\
Foxa2 & 2.70 &  Anterior Visceral endoderm, Anterior primitive streak & \cite{otx2}, \cite{foxa2PS} \\
Gsc & 2.67 & Anterior Primitive Streak & \cite{Lewis:2007aa} \\
Hand1 & 2.57  &Posterior Mesoderm, Lateral Plate Mesoderm & \cite{Riley:1998aa} \\
Ascl2 & 2.52 & Ectoplacental Cone &\cite{Simmons:2005aa} \\
Mesp1 & 2.46  & Posterior Primitive Streak & \cite{Arnold_2009} \\
Hoxa1 & 2.44  & Mesoderm & \cite{hox} \\
Nanog & 2.24 & Epiblast & \cite{Mulas:2018aa} \\
Zfp42 & 2.14  &Extraembryonic Ectoderm & \cite{Pelton329} \\
Cdx1 & 2.12  & Paraxial Mesoderm & \cite{vandenAkker:2002aa} \\
Runx1 & 1.82 & & \\
Hoxb2 & 1.73 & Mesoderm & \cite{hox} \\
Id2 & 1.51 & Extraembryonic Ectoderm & \cite{id2} \\
Tbx3 & 1.31  & &\\
Pitx2 & 1.20  & &\\
 \hline
\end{tabular}

We attempted to discover a low dimensional subspace in which signatures of cell type diversity could be inferred using the algorithm outlined in the previous section. We sampled 3000 clustering configurations based on hierarchical (ward) clustering of 5000 subsampled cells, with with $K_{p} \sim Unif(20,75)$, chosen to cover a range around the 37 clusters found in \cite{gottgens}. We find 27 transcription factors with a z-score $z_g > 1$,  18 of which have known have previously identified essential functions in the regulation of differentiation during gastrulation (See Table \ref{table1}). \\

Our hypothesis is that the variation in the 27 discovered transcription factors provides a subspace $V_s$ in which multimodal signatures allow the identification of cell types. However, single cell measurements of individual genes are known to be subject to a variety of sources of technical noise \cite{Gr_n_2015}. In order to decrease reliance on individual measurements, we take each of the 27 transcription factors with high scores, and extend the subspace to include 5 genes (potentially not transcription factors) that have the highest correlation with each of the 27 discovered transcription factors, resulting in an expanded subspace of 83 genes in which to cluster the data (Full list in supplement). The cell-cell covariance matrix in this subspace (Fig. \ref{fig5}B), reveals distinct cell types and subtypes, and a heat map of the expression levels of these 83 genes shows differential expression between subtypes of cells.

\begin{tabular}{ p{1.2cm}||p{5cm}|p{5cm}  }
 \multicolumn{3}{c}{Table 2} \\
 \hline
Cluster ID & Label &  Markers\\
\hline
0 & Unclustered & \\
1 & Primitive Erythrocyte Progenitor \cite{gata1} & Hba-x \cite{hbax}, Hbb-bh1 \cite{hbbbh1}, Gata1 \cite{gata1}, Lmo2 \cite{lmo2} \\
2 & Mesoderm &  Car3, Spag5, Hoxb1 \cite{hox} , Cnksr3, Smad4, Zfp280d, Vim \cite{VIM}, Ifitm1 \\
3, 4 & Lateral Plate Mesoderm & Foxf1 \cite{Mahlapuu:2001aa}, Hand1 \cite{Riley:1998aa} \\
5 & Anterior Visceral Endoderm & Sox17, Foxa2 \cite{otx2}, Cer1 \cite{cer1}, Frat2, Lhx1 \cite{lhx1}, Hhex \cite{hhex}, Gata6, Ovol2, Otx2 \cite{otx2}, Sfrp1 \cite{sfrp1} \\
6,7 & Extraembryonic Mesoderm & Fgf3 \cite{Niswander755}, Lmo2 \cite{lmo2}, Gata2 \cite{gata2}, Bmp4 \cite{bmp4} \\
8,9 & Visceral Endoderm 1 &  Rhox5 \cite{rhox5}, Emb \cite{emb}, Afp \cite{afp} \\
10,11 & Neuromesodermal Progenitor & Sox2, T \cite{NMP} \\
12,13& Paraxial Mesoderm / Presomitic Mesoderm & Hoxa1, Hoxb1 \cite{hox}, Cdx1, Cdx2 \cite{vandenAkker:2002aa} \\
14 & Posterior Primitive Streak & Mesp1 \cite{Arnold_2009}, Snai1 \cite{snai1}, Lhx1 \cite{lhx1, lhx1b}, Smad1 \cite{smad1} \\
15 & Anterior Primitive Streak, Organizer-like Cells & Foxa2 \cite{foxa2PS}, Gsc \cite{Lewis:2007aa}, Eomes \cite{foxa2PS} \\
16,17 & Posterior Primitive Streak derived Mesoderm, Lateral Plate Mesoderm Progenitors & Msx2 \cite{msx2}, Snai1 \cite{snai1}, Foxf1 \cite{Mahlapuu:2001aa}, Hand1 \cite{Riley:1998aa}, Gata4 \cite{gata4}\\
18,19 & Extraembryonic Ectoderm & Cdx2 \cite{cdx2}, Rhox5 \cite{rhox5}, Id2 \cite{id2}, Gjb5 \cite{Frankenberg_2007}, Tfap2c \cite{Latos_2015}, Zfp42(aka Rex1) \cite{Pelton329}, Elf5 \cite{Latos_2015}, Gjb3 \cite{Frankenberg_2007}, Ets2 \cite{DONNISON201577}\\
20 & Ectoplacental Cone &  Plac1 \cite{DONNISON201577}, Ascl2 \cite{Simmons:2005aa}\\
21 & Definitive Endoderm & Sox17 \cite{Viotti:2014aa}, Foxa2 \cite{Burtscher:2009aa}, Apela \cite{Hassan_2010}\\
22 & Mesendo Progenitor, Primitive Streak & Tcf15 \cite{Chal:2018aa}, Cer1, Hhex \cite{Thomas:1998aa} \\
23 & Posterior Primitive Streak, Cardiac Mesoderm Progenitors & Mesp1 \cite{Arnold_2009}, Gata4 \cite{gata4}, Lhx1 \cite{lhx1b}, Smad1 \cite{smad1}\\
24,25 & Primitive Streak & T, Mixl1, Eomes, Fgf8, Wnt3\\
26 & & Klf10, Gpbp1l1, Hmg20a, Rbm15b, Celf2 \\
27-28 & Posterior-Proximal Epiblast & Nanog \cite{Mulas:2018aa}, Sox2 \cite{Avilion_2003}, Pou5f1 \cite{Mulas:2018aa}, Otx2 \cite{Kurokawa:2004aa}\\
29-34 & Epiblast & Sox2  \cite{Avilion_2003}, Pou5f1 \cite{Mulas:2018aa}, Otx2 \cite{Kurokawa:2004aa} \\
\hline
\end{tabular}

We hierarchically clustered the cells into 35 cell types based on expression of these 83 genes. The corresponding identity of these cell types was determined using the expression pattern of all genes (Table 2, Fig. \ref{fig5} C), and identify extraembryonic populations (C5-9,C18-20), epiblast populations (C27-C34), primitive streak populations (C14-17,C23-25), mesoderm subtypes (C2-4,C12,C13,C16,C17,C22), endoderm (C21), and primitive erythrocyte (C1). For example, we find a subpopulation of epiblast cells that have upregulated Nanog (as well as other early markers of the primitive streak), suggesting that these cells are positioned on the posterior-proximal end of the epiblast cup \cite{Mulas:2018aa}. The large primitive streak population, which extends along the proximal side of the embryo, contains subtypes distinguished by Gsc \cite{Lewis:2007aa} and Mesp1 \cite{Arnold_2009}, which give rise to distinct fates. We find a distinct population of anterior visceral endoderm cells, marked by Otx2 and Hhex, which define the population responsible for the anterior-posterior body axis \cite{otx2}. This population, which is distinguished from other Foxa2-expressing subpopulations of the visceral endoderm, is crucial for proper development.

Most importantly, in extracting the relevant features from the data, our algorithm identifies known and validated transcription factors that are crucial to the developmental processes happening in this time frame. Further, by eliminating extraneous measurements, we are able to identify clear differential expression patterns between sub-types of cells which were indistinguishable through previous methods. In particular, identification of primitive streak subpopulations provides novel insight into a central developmental process, and we identify key genes that would allow for experimental interrogation of the spatial organization of the sub-types and their dynamics.\\

\subsection{Comparison to Existing Methods}\label{exmm}

There are three main classes of methods for feature identification. The first class builds on correlation analysis and PCA. Of these, most methods operate in the full space $V$ \cite{correlatedfeatures}. These methods fail as $D/D_s$ becomes large, which is the regime of interest in this paper (See Supplemental Fig. 1). This has been addressed through Sparse PCA \cite{SPCA_2009}, which is effective in finding sparse representations of the data. Sparse PCA has two main drawbacks: 1) analysis depends on the choice of two free parameters (number of principal components chosen, and a sparsity parameter. See Supplemental Fig. 3, top row), which makes it difficult to identify individual features of importance, 2) it does not optimize for any notion of data separability.

The second class of methods are model based. The identification of a small subset of informative features can be formulated as a Bayesian inference problem, where a log likelihood function is maximized over the hidden parameters via an expectation maximization scheme \cite{dempster1977}. Model based clustering has been explored in depth in \cite{MCLA2000, FraleyRaferty}, and adapted to feature selection by the inclusion of a lasso term on the separation of the first moments in \cite{PanShen07,Wang_2008,Xie_2008}. These methods all rely on accurate forward models of the data. Advantages and drawbacks of these are discussed in \cite{skmeans}, which provides a more general framework.

The third class of methods for feature detection are based on clustering. In \cite{LFSBSS}, features are discovered based on ability to define individual clusters, but this method cannot resolve distinct clusters in the large $D/D_s $ limit (Supplemental Fig. 3D). In the final method in this class \cite{skmeans}, a feature weight vector $\vec{w} \in \mathbb{R}^D$ is introduced and learned by amending a clustering cost function with a $L1$ penalty on the feature weights. This optimization proceeds somewhat differently depending on the underlying clustering algorithm in use (e.g. K-Means, K-Medoids, Hierarchical clustering). The K-Means version is successful in discovering structure in the large $D/D_s$ regime with Gaussian data (Supplemental Fig. 3E), yet the hierarchical clustering procedure does not do as well (Supplemental Fig. 3F). Additionally, in situations with a large number of data points $N$, the hierarchical clustering approach requires the construction of a $N^2 x D$ matrix, which is computationally difficult. Further, these methods both rely on knowing the number of clusters, which is an input to the algorithm, and is difficult to infer \cite{gapstat_talk}. Therefor, in situations with real data when K-Means is a poor choice of a clustering algorithm (non-spherical data, uneven cluster size), the computational inefficiencies of sparse hierarchical clustering are limiting.

Our approach has a number of general advantages. First, it does not make assumptions about the number of clusters, the types of generating distributions, or the relative sizes of the different clusters. Second, by integrating over an ensemble of proposal cluster configurations constructed on subsets of the data, the algorithm is computationally efficient in regimes of large $N$ (does not suffer from the $N^2$ scaling of sparse hierarchical clustering). Third, by building on existing clustering methods to construct proposals, our method can be generally applied over any clustering procedure to discover relevant features.

\section{Discussion}

Identifying sub-spaces which define classes and states from high dimensional data is an emerging problem in scientific data analysis where an increasing number of measurements push the limits of conventional statistical methods. Techniques such as PCA and ICA provide invaluable insight in data analysis, but can miss multimodal features, particularly in high dimensional settings. These methods which have reduced success in the $D/D_s >> 1$ regime can be supplemented by our technique by finding a lower dimensional subspace in which further analysis can be conducted. Crucially, eliminating any informative dimensions decreases the $D/D_s$ ratio, moving to a regime in which conventional methods are more effective. By reducing the dimensionality of the data, it is possible to artificially increase data density, and mitigate associated problems that are prevalent in high dimensional inference. Further, as our algorithm can be a wrapper over any clustering algorithm to construct the proposal clusters, it has varied applicability in settings where K-means or other specific clustering algorithms are unsuccessful. 

Biological data from neural recordings, behavioral studies, or gene expression is increasingly high dimensional. Identifying the underlying constituents of the system that define distinct states is crucial in each setting. In contexts such as transcriptional analysis in developmental biology, finding the key genes that define cell states is a central problem that bridges the gap between high throughput measurements and mechanistic experimental follow ups. Identification of transcription factors with multimodal expression that define cellular states allows for the study of dynamics of state transitions and spatial patterning of the embryo. Our method rediscovers known factors in well studied developmental processes and predicts several gene candidates for further study. Identifying defining features in high dimensional data is a crucial step in understanding and experimentally perturbing systems in a range of biological domains.

\section*{Acknowledgements}
We would like to thank Deniz Aksel for extensive help with annotating the clusters. In addition we would like to thank Cengiz Pehlevan, Sam Kou, Matt Thomson, Gautam Reddy, Sean Eddy, and members of the Ramanathan lab for discussions and comments on the manuscript. The work was supported by DARPA W911NF-19-2-0018 , 1R01GM131105-01 and DMS-1764269.

\bibliographystyle{unsrt}  
\bibliography{library}

\newpage
\section{Supplemental Information}\label{app}

\beginsupplement

\begin{figure}[h!]
\includegraphics[width=\textwidth]{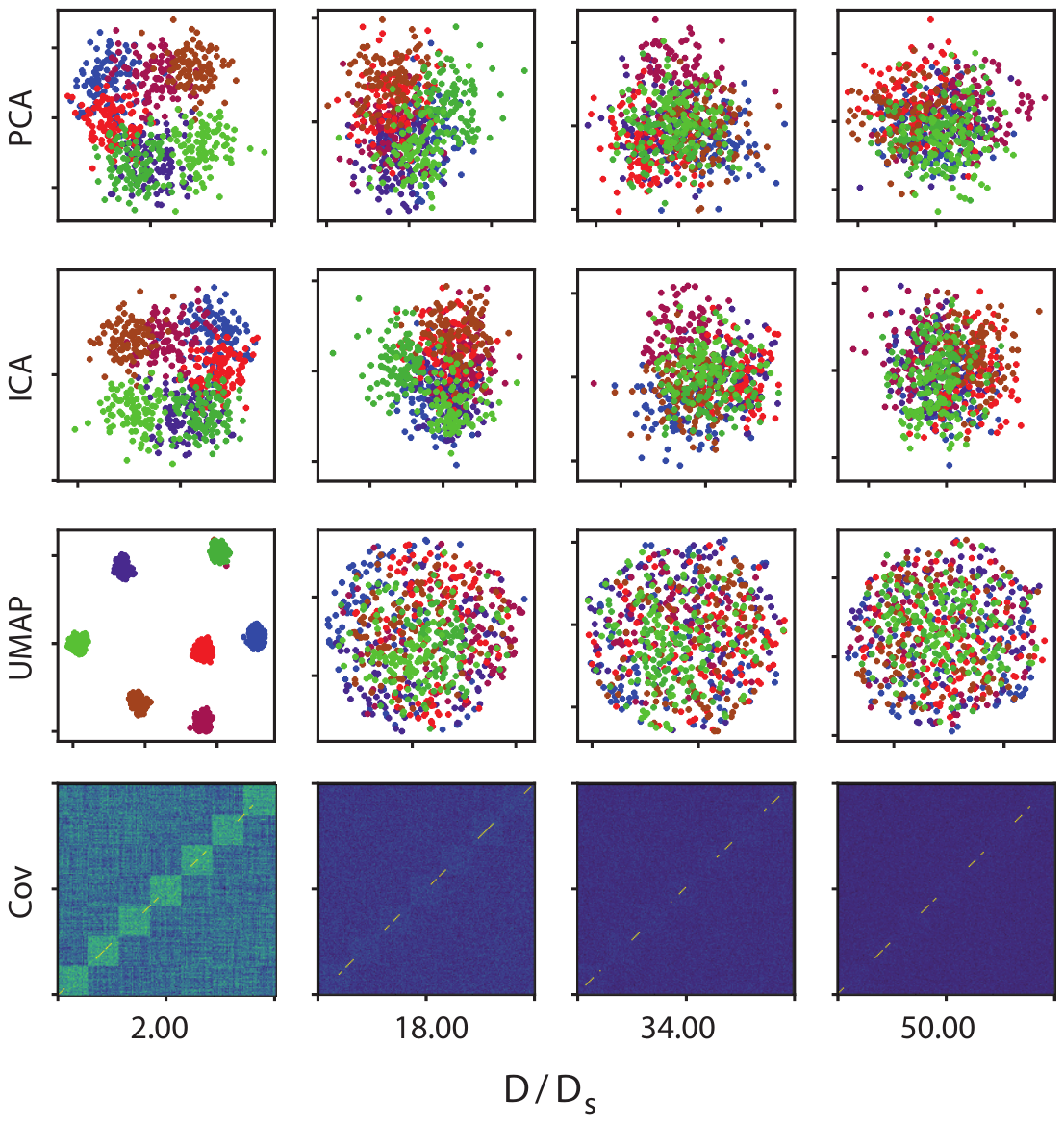}

\caption{As the ratio of the number of informative dimensions (in which clusters are linearly separable) to the number of uninformative or noisy dimensions increases, conventional dimensionality reduction methods fail to separate clusters (colors). Here 1400 data points are generated from a Gaussian mixture, where the means and variances of each cluster are identical in every dimension except for a subspace of dimension $D_s = 21$. In this subspace, each cluster is linearly separable from all other clusters. Rows correspond to common visualization and dimensionality reduction methods, which fail to separate distinct multimodal clusters as $D/D_s$ increases. Crucially, techniques that build on top of PCA or covariation analysis all rely on extraction of signal through these methods.}
\label{figS1} 
\end{figure}

\begin{figure}

\includegraphics[width=\textwidth]{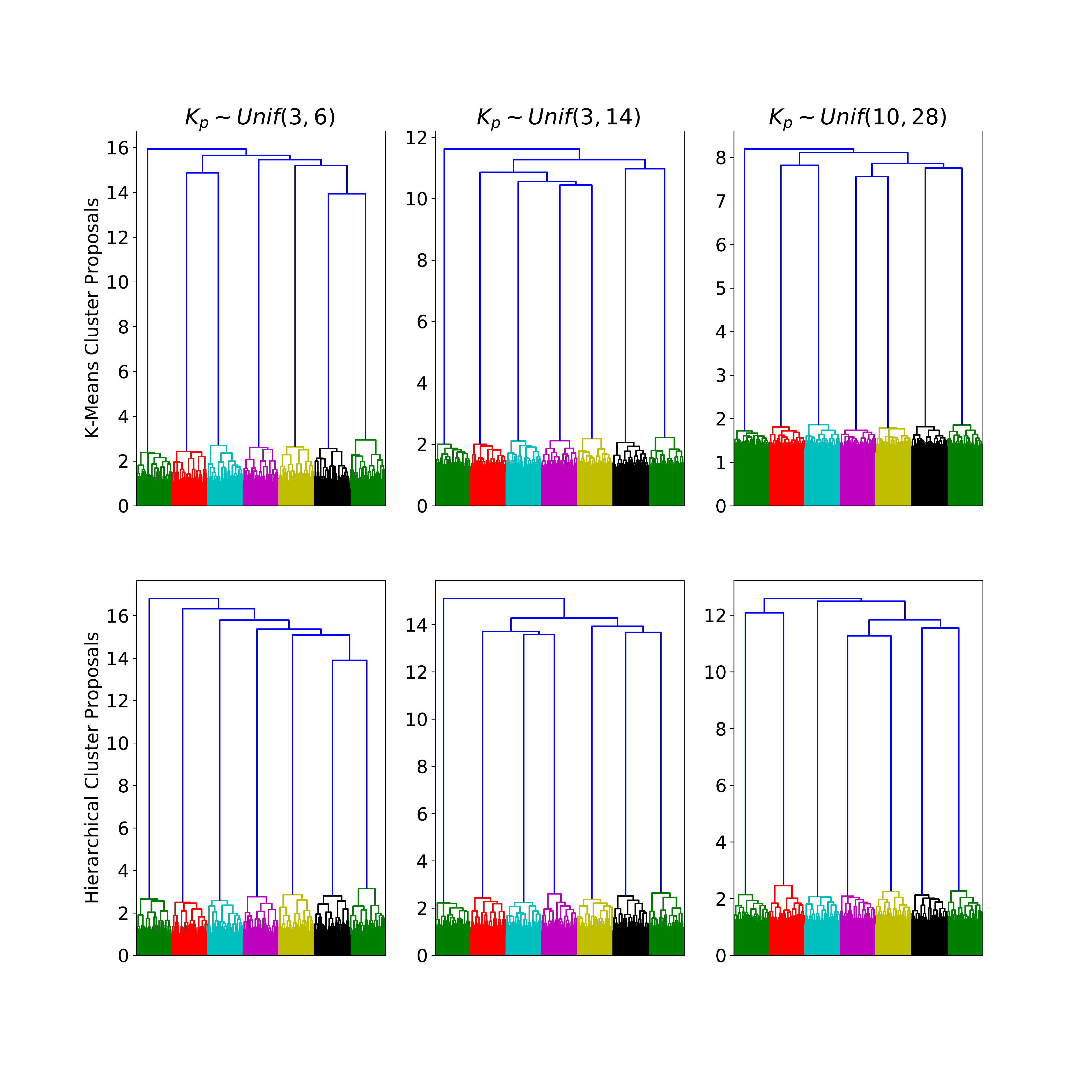}

\caption{The algorithm reliably discovers a set of features in which the existence of 7 clusters is revealed. Here 200 points from 7 distinct clusters were generated from Gaussian distributions where 21 features are pairwise informative as to cluster identity, and 840 features are uninformative (thus $D/D_s = 40$). See main text for details. Using either K-Means (top row) or hierarchical clustering (bottom row) to construct proposal clusters results in a subspace in which separation between clusters is visible. Different priors from which $K_p$ is drawn are shown in the different columns, but do not greatly effect performance. See Main Text Fig. 3A for comparison of a dendrogram constructed in the full space.}

\label{figS2} 
\end{figure}

\begin{figure}

\includegraphics[width=\textwidth]{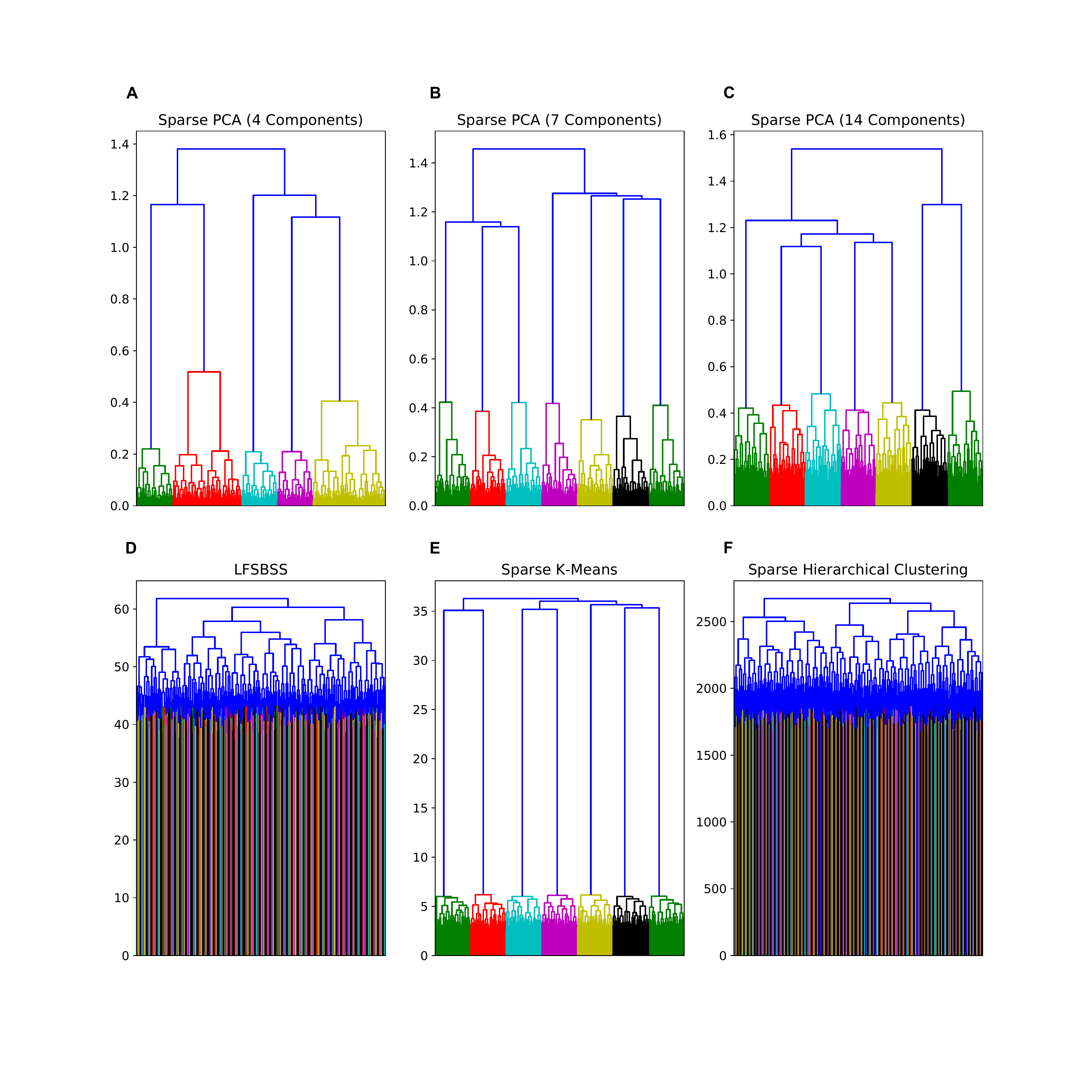}

\caption{Comparison to existing methods, all done on Gaussian data with 7 clusters and $D/D_s = 40$. A)-C) Sparse PCA finds useful representations, but depends on a sparsity parameter and the representation is dependent on how many components are kept. D) LFSBSS attempts to construct locally relevant feature weights, but does not identify relevant clusters in this $D/D_s$ regime. E) Sparse K-means discovers the true clusters and features. D) Sparse hierarchical clustering is computationally ineffecient, and fails to resolve the correct features in this setting.}

\label{figS3} 
\end{figure}

\newpage

\subsection{Correlations and Dimensionality}\label{appA}

In situations where some dimensions or features do not carry relevant information 
For data points $\vec{x} \in \mathbb{R}^D$, if $V \in \mathbb{R}^{D_1}$ is the subspace with meaningful information, and $U \in \mathbb{R}^{D_2}$ is just noise.

We are first interested in distances between $\vec{x}_1, \vec{x}_2 \in \mathbb{R}^D$. Let $\vec{x}^V = \vec{x} \bot V$. We first want to compute the correlation between $| \vec{x}^V - \vec{x}_2^V |^2$ and $| \vec{x}_1 - \vec{x}_2 |^2$. If $A = \vec{x}_1^V - \vec{x}_2^V$ and $B = \vec{x}_1^U - \vec{x}_2^U$ then:

$$c =  \frac{\langle \left( A^2 -  \langle A ^2 \rangle \right) \left( A^2 + B^2 - \langle A^2 + B^2 \rangle \right) \rangle}{\sigma_{A^2} \sigma_{A^2+B^2}}$$ 
$$ = \frac{1}{\sigma_{A^2} \sigma_{A^2+B^2}} \left( \langle A^4 \rangle  + \langle A^2 B^2 \rangle - \langle A^2 \rangle ^2 - \langle A^2 \rangle \langle B^2 \rangle \right)$$

By definition, $\langle A^4 \rangle  - \langle A^2 \rangle ^2 = \sigma^2_{A^2}$, and so $$ \frac{\sigma_{A^2} }{\sigma_{A^2 + B^2}} + \frac{1}{\sigma_{A^2} \sigma_{A^2+B^2}}\left( \langle A^2 B^2 \rangle - \langle A^2 \rangle \langle B^2 \rangle \right) $$ The signal in $V$ and $U$ is uncorrelated, so the second term is $0$. $A, B$ are both differences of multivariate Gaussians, which are also multivariate Gaussian. And if we let $C = \vec{x}_1-\vec{x}_2$ then $C$ is also multivariate Gaussian. We want to compute $\sigma_{A^2}/\sigma_{C^2}$. For a multivariate Gaussian $Z \sim \mathcal{N}(0,\Sigma)$, $\langle Z^2 \rangle  = Tr(\Sigma)$,  $\langle Z^4 \rangle  = 2 Tr(\Sigma^2) + Tr(\Sigma)^2$

$$\sigma_{A^2} = \left( \langle A^4 \rangle - \langle A^2 \rangle^2\right)^{1/2} = \left( 2Tr(\Sigma^2_V) \right)^{1/2}$$
$$\sigma_{C^2} =  \left(  \langle C^4 \rangle - \langle C^2 \rangle^2 \right)^{1/2}=  \left( 2Tr(\Sigma^2) \right)^{1/2}$$

So $$ c= \left[\frac{Tr(\Sigma^2_V)}{Tr(\Sigma^2) }\right]^{1/2}$$

The trace is the sum of the eigenvalues squared, so if the data is normalized such that the eigenvalues are of order 1, this scales like $$ c \sim \left( \frac{D_1}{D_1 + D_2} \right)^{1/2}$$

\subsection{False Positive and False Negative Rates}\label{appB}

  Let $m_i$ be the number of proposed clusters per true cluster ($i \in [1,K_{true}]$). We assume that the entropy relative to the true class labels is low in each of the proposed clusters (in practice, this condition is met in situations where $K_{p} > K_{true}$). This is true as long as the prior on the proposal cluster number $P(K^{p})$ is non-zero for $K^{p}>K^{true}$, because the expectation of $g_d$ in Equation \ref{main_eqn_end} is dominated by larger values of $K^p$ because each proposal contributes $\sim K^{p2}/2$ terms to the sum. We start by counting the number of pairs of proposed clusters wherein cluster in the pair contains data points primarily belonging to the same true cluster:
  
  $$\sum_{i=1}^{K_{true}} \binom{m_i}{2}$$

And so the frequencies of errors is given by $$f_e = \frac{\sum_{i=1}^{K_{true}} \binom{m_i}{2}}{\binom{K_{p}}{2} \left( D - \binom{K_{true}}{2}\right)}$$

If the true clusters have roughly equal sizes, then we can assume that $m_i \approx m_j \approx K_{p}/K$ for all $i,j$, and this ratio reduces to $$f_e=\frac{K_{true}}{D-\binom{K_{true}}{2}}  \frac{\binom{K_{p}/K_{true}}{2}}{\binom{K_{p}}{2}}$$

The signal frequency is given by a count for each feature when the two proposal clusters in a pair contain points primarily from different true clusters, of which there are roughly $$\sum_i \left[ m_i\cdot \left( \sum_{j \neq i} m_j\right)\right]/2 $$

These votes all identify an informative feature, so the fraction is given by 
$$f_s = \frac{\sum_i \left[ m_i\cdot \left( \sum_{j \neq i} m_j\right)\right]/2 }{\binom{K_{p}}{2}}$$

When we assume each $m_i \approx \frac{K_{p}}{K_{true}}$, this reduces to: $$f_s \approx \frac{1}{2}\left( \frac{K_{p}}{K_{true}}\right)^2\left(\frac{1}{\binom{K_{p}}{2}}\right)$$

The ratio of $f_s/f_e$ tells us the regime in which the noise is much lower than the signal for meaningful features:

$$ \frac{f_s}{f_e} \approx \frac{\left( \frac{K_{p}}{K_{true}}\right)^2\left(D-\binom{K_{true}}{2}\right)}{2 K_{true} \binom{K_{p}/K_{true}}{2}}$$

$$ = \frac{K_{p}^2 \left( D - \binom{K_{true}}{2} \right) }{2 K_{true}^3 \left( \frac{K_{p}}{K_{true}} \right)  \left( \frac{K_{p}}{K_{true}}-1 \right)}$$

$$ \approx \frac{K_{p}^2 \left( D - \binom{K_{true}}{2} \right) }{2 K_{true}^3 \left( \frac{K_{p}}{K_{true}} \right)^2}$$

$$ \approx \frac{ D - \binom{K_{true}}{2}  }{2 K_{true}}$$

Here $ D - \binom{K_{true}}{2} $ are the number of uninformative features, and $D_1 \propto K_{true}^2$ is the number of informative features, so this scales like $$ \approx \frac{D-D_s}{D_s^{1/2}} \approx \frac{D}{\sqrt{D_s}}$$

\end{document}